\documentclass[10pt,journal]{IEEEtranTCOM}
\usepackage[english]{babel}
\usepackage[usenames]{color}
\usepackage[cp1250]{inputenc}
\usepackage{amsfonts}
\usepackage{amsthm}
\usepackage{graphicx}
\usepackage{epsfig}
\usepackage{mathrsfs}
\usepackage{amsmath}
\usepackage{algorithm}
\usepackage{algorithmic}
\usepackage{hyperref}

\theoremstyle{plain}

\begin{document}
\newcommand{\bea}{\begin{eqnarray}}
\newcommand{\eea}{\end{eqnarray}}
\newcommand{\be}{\begin{equation}}
\newcommand{\ee}{\end{equation}}
\newcommand{\beas}{\begin{eqnarray*}}
\newcommand{\eeas}{\end{eqnarray*}}
\newcommand{\bs}{\backslash}
\newcommand{\bc}{\begin{center}}
\newcommand{\ec}{\end{center}}
\def\SC {\mathscr{C}}

\title{Predicting conditional probability distributions \\of redshifts of Active Galactic Nuclei \\ using Hierarchical Correlation Reconstruction }
\author{\IEEEauthorblockN{Jarek Duda}\\
\IEEEauthorblockA{Jagiellonian University,
Golebia 24, 31-007 Krakow, Poland,
Email: \emph{dudajar@gmail.com}}}
\maketitle

\begin{abstract}
While there is a general focus on prediction of values, real data often only allows to predict conditional probability distributions, with capabilities bounded by conditional entropy $H(Y|X)$. If additionally estimating uncertainty, we can treat a predicted value as the center of Gaussian of Laplace distribution - idealization which can be far from complex conditional distributions of real data. This article applies Hierarchical Correlation Reconstruction (HCR) approach to inexpensively predict quite complex conditional probability distributions (e.g. multimodal): by independent MSE estimation of multiple moment-like parameters, which allow to reconstruct the conditional distribution. Using linear regression for this purpose, we get interpretable models: with coefficients describing contributions of features to conditional moments. This article extends on the original approach especially by using Canonical Correlation Analysis (CCA) for feature optimization and l1 "lasso" regularization, focusing on practical problem of prediction of redshift of Active Galactic Nuclei (AGN) based on Fourth Fermi-LAT Data Release 2 (4LAC) dataset.
\end{abstract}
\section{Introduction}
Machine learning is usually focused on prediction of values. If additionally estimating uncertainty, we get (unimodal) conditional distributions e.g. as Gaussian or Laplace - it is crucial to do it in data compression, but generally it is relatively uncommon practice.

\begin{figure}[t!]
    \centering
        \includegraphics[width=8.5cm]{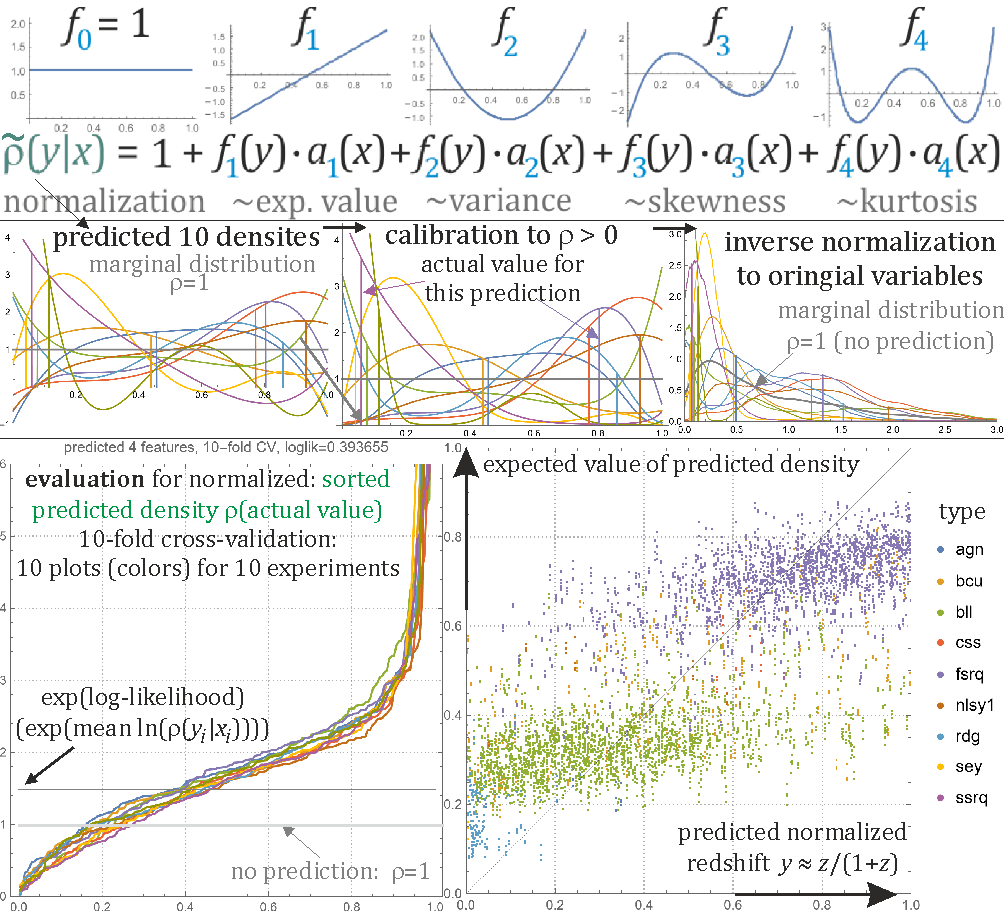}
        \caption{Top: for predicted variable normalized to nearly uniform distribution ($y\approx z/(1+z)$ for redshift), there is modeled predicted conditional density as linear combination of moment-like orthonormal functions (further optimized with CCA), here using least-square linear regression of features of $k=21$ variables. Center: such predicted linear combination might get below zero, requiring to use calibration $\rho=\phi(\tilde{\rho})/Z$ (here using optimized softmax-like: $\phi(\rho)=\ln(1+\exp(3\rho)/4)/3$) and then normalizing to integrate to 1. Finally we can inverse the normalization to return to the original variable (redshift here). Bottom left: evaluation using 10-fold cross-validation, there are presented sorted predicted densities in the actual values, and log-likelihood averaging them, which is estimated conditional entropy $-H(Y|X)$. As low probability values also have a chance to happen, in $\approx 20\%$ of cases it is below no prediction $\rho=1$. Bottom right: actual normalized values (horizontal) vs expected values for their predicted densities (vertical) - which can be treated as cautious predicted values: avoiding extremes. Fig. \ref{mv} shows means with variances. }
        \label{intr}
\end{figure}

\begin{figure*}[t!]
    \centering
        \includegraphics[width=18cm]{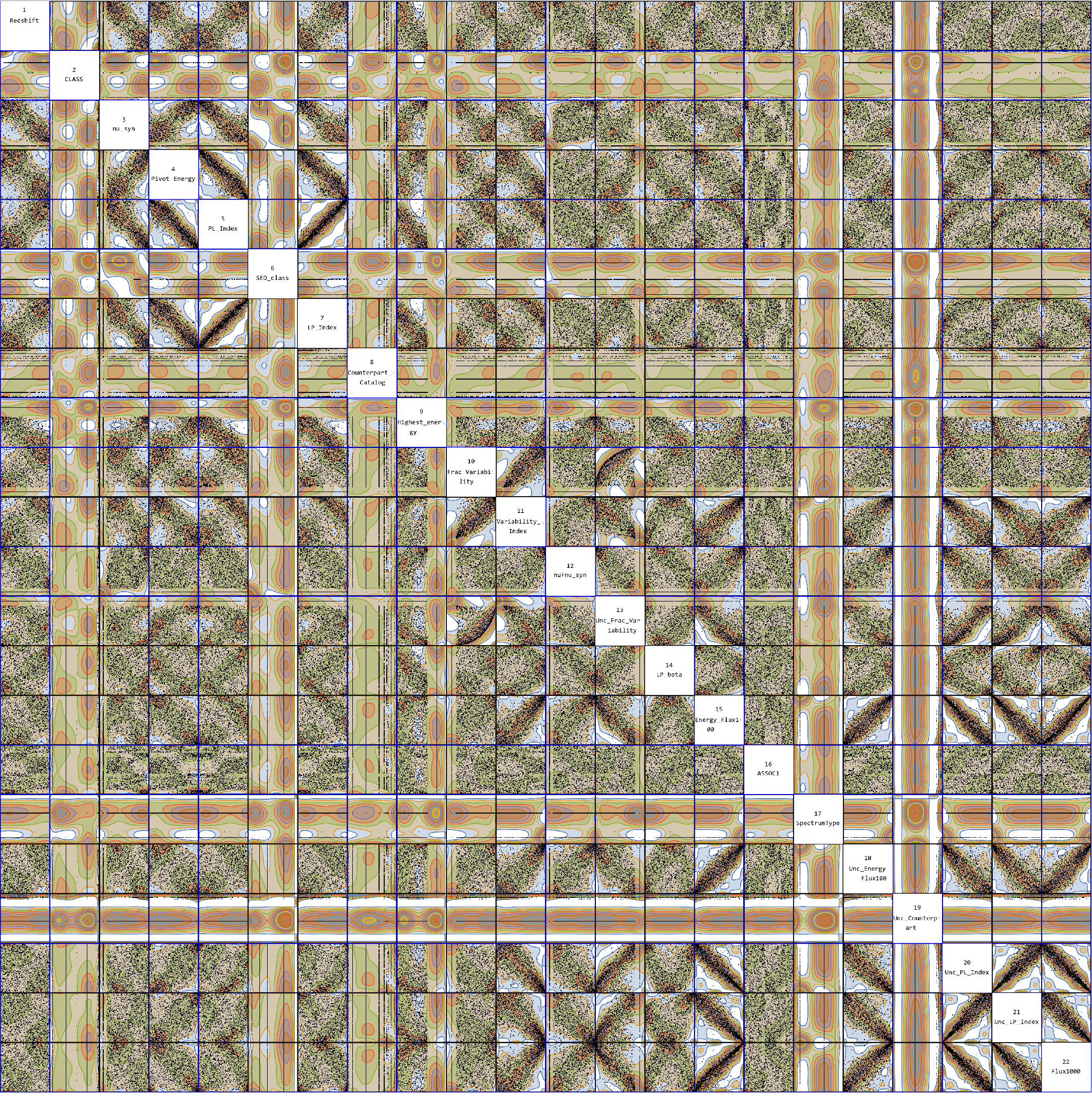}
        \caption{Used 22 variables, each normalized to nearly uniform distribution in $[0,1]$ using empirical distribution function - shown in pairs (1767 available points) with visualized joint distribution estimated using degree 4 polynomials (shown isolines for $\rho=0,0.5,1,1.5,2,\ldots$). We want to predict conditional distribution of the first (redshift) from the remaining 21. They are sorted by log-likelihood (estimated $-H(Y|X_j)$), shown in Fig. \ref{rn}) if predicting from a given single variable. While most of variables are continuous, some have lines representing identical values - discrete variables, or e.g. missing values for mostly continuous variables.
        We can see that conditional distributions are quite complex already for pair-wise dependencies, hence just prediction of value is not a good idea here.}
        \label{pairs}
\end{figure*}

However, for real data the conditional distributions are often much more complex, e.g. in Fig. \ref{pairs} we can see pairwise joint distributions for discussed here astronomical data. The prediction capabilities are bounded by conditional entropy $H(Y|X_j)$.  We would like to predict conditional distribution from multiple variables (bounded by $H(Y|X_1..X_k)$), what is much more difficult - here from the looking most valuable $k=21$ variables, which are of various types: discrete, continuous, or combined - mostly continuous, but also with discrete part: here it is missing value or 0, 10 values.

Advantages of probability distribution prediction are, among others: uncertainty estimation (also of skewness, kurtosis), credibility evaluation - values in low density might be interesting or worth reexamination, allows to apply further nonlinear functions $(E[g(X))]\neq g(E[X]))$, or generate synthetic data with such distribution e.g. for Monte Carlo.

This article expands on Hierarchical Correlation Reconstruction (HCR) approach for this purpose - discussed more deeply in \cite{hcr1,hcr2}, briefly presented in Fig. \ref{intr}. Specifically, we first normalize the predicted variable $Y$ to nearly uniform distribution on [0,1] as in copula theory~\cite{copula}, then predict this conditional distribution as a linear combination - using orthonormal $f_i$ basis:  $\tilde{\rho}(Y=y|X=x)=\sum_i a_i(x)f_i(y)$. Its $(a_i)$ parameters can be estimated from the remaining variables $(x_j)$ independently with MSE (mean-squared estimation). While density has to be nonnegative, such linear combination can get below zero - requiring to further apply calibration $\rho=\phi(\tilde{\rho})/Z$ for e.g. for $\phi(x)=\max (x,0.1)$, and $Z$ normalization to integrate to 1. Then we can go back through normalization of $y$, getting predicted density of the original variable.

The $(a_i)$ moment-like parameters of predicted conditional distributions could be MSE estimated from $(x_j)$ variables using various techniques up to neural networks. For interpretability we will focus on linear regression - of also moment-like features of $(x_j)$, could be also including their products for multi-variate dependencies.

The main difficulty is rapid growth of size of such models especially if including multi-variate dependencies. Hence optimization requires subtle feature selection and regularization - this article extends on: using Canonical Correlation Analysis (CCA) which allows to optimize the predicted basis and used features of $(x_j)$ variables, and l1 regularization "lasso" allowing to find smaller sparse models preventing overfitting. 

Finally the models were evaluated with log-likelihood for 10-fold cross-validation: splitting datastet into 10 subsetes, one is used as test, the remaining for training in all 10 ways. Log-likelihood evaluation averages over such 10 experiments.

In this article we focus on Fourth Fermi-LAT Data Release 2 (4LAC) data~(\cite{dat1,dat2}) of Active Galactic Nuclei (AGN)  described with 38 continuous and/or discrete variables. One of them is redshift, which is difficult to measure experimentally, hence there is active research to predict it from the remaining variables - currently focused on prediction of values~(\cite{ml1,ml2}). Proposed here prediction of probability distribution can shows additional dependencies, suggests less credible values for examination, can be used as prior distribution for some further analysis, and so on.

This is initial version of article, planned to be improved - through consultation with expert in this field e.g. for practical selection of variables and possible applications. Also to try to improve the methodology, especially feature selection and regularization, maybe try stronger models to predict $(a_i)$ parameters like neural networks.

\section{Data and analysis}
\begin{figure}[t!]
    \centering
        \includegraphics[width=8.5cm]{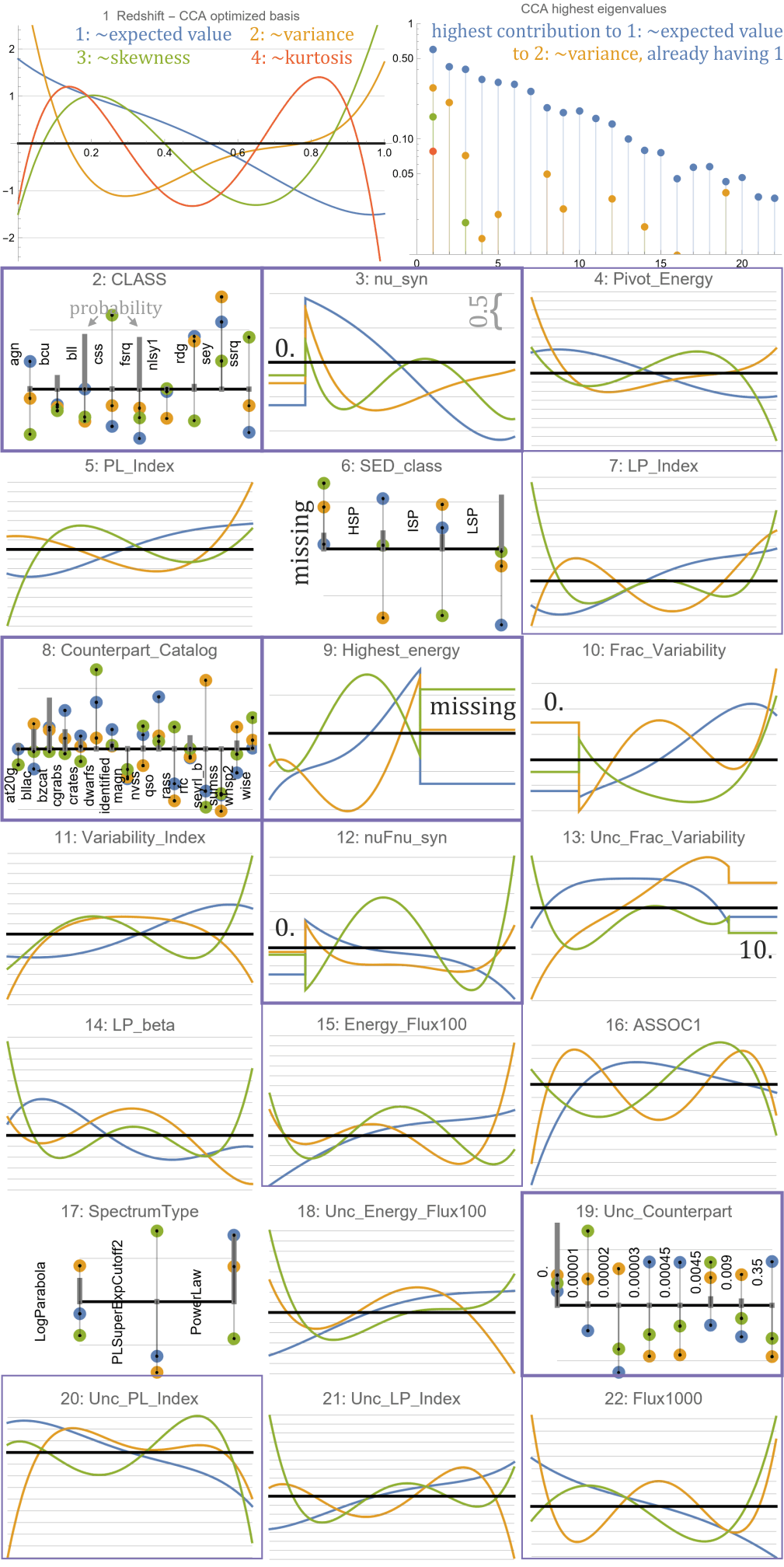}
        \caption{CCA-optimized features of the 22 variables - presented first 4 or 3 with correspondingly: blue, orange, green, red colors. Marked variables are the most valuable from perspective of novelty analysis in Fig. \ref{rn} - carrying valuable information unavailable in the remaining. Top right plot shows the highest eigenvalues of obtained cross-covariance matrix, estimating importance of such features - as alternative for relevance in Fig. \ref{rn}.  }
        \label{vars}
\end{figure}

\begin{figure}[t!]
    \centering
        \includegraphics[width=8.5cm]{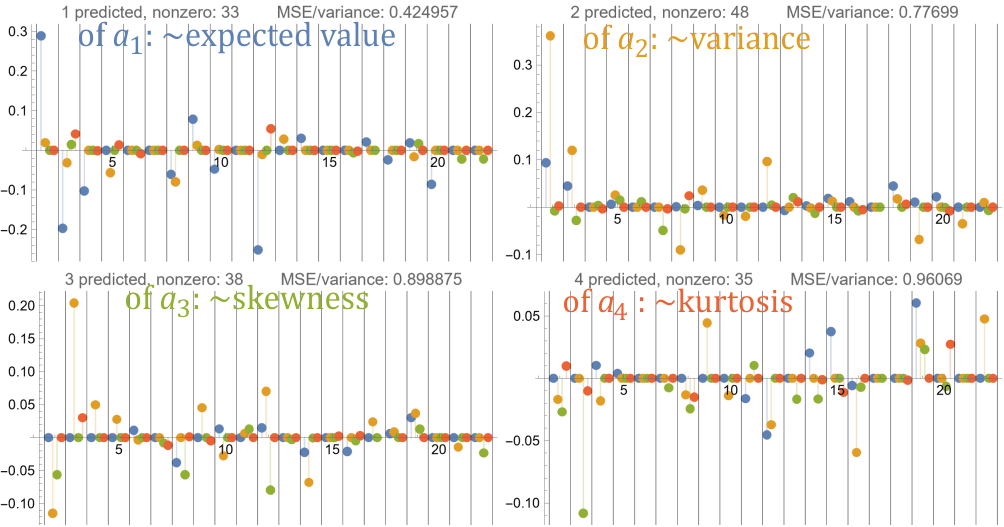}
        \caption{There was used least-squares linear regression with l1 (lasso) regularization to predict $a_1,a_2,a_3,a_4$ coefficients in optimized basis presented in Fig. \ref{vars}, based on features of the 21 variables. The shown obtained sparse nonzero coefficients can be interpreted as contributions of moments of 21 variables to moments of the predicted variable. For $a_1$ (top-left) mostly the first features (blue) - due to CCA optimization. }
        \label{lasso}
\end{figure}
\subsection{Data source and variable selection}
The Fourth Fermi-LAT Data Release 2 (4LAC) dataset  was downloaded from \url{https://fermi.gsfc.nasa.gov/ssc/data/access/lat/4LACDR2/}: two files for low and high latitude, containing 37 variables for 380 + 3131 = 3511 objects. As we are interested in prediction of redshift, there were selected all 1767 objects having this value (it is missing for the remaining), also datasets from both files were merged - treating latitude as additional 38th variable.

Then with analogous  simplified methodology there were estimated log-likelihoods for prediction of redshift from each individual variable (relevance in Fig. \ref{rn}). For simplicity there were left only those above 0.01 - leaving 21 used variables, sorted by this evaluation.
\subsection{Variable normalization to nearly uniform distributions}
In the discussed methodology, analogously to copula theory~\cite{copula}, it is convenient to predict variables normalized to nearly uniform distribution on $[0,1]$. For redshift $z$ there could be used standard $y=z/(1+z)$ transformation for this purpose.

However, a bit better performance provides normalization with empirical distribution, so it is used here also for the remaining variables. Finally the predicted density can be taken to the original variable by inverting this normalization.

Specifically, for normalization with empirical distribution, all $n$ values are first sorted (alphabetically for non-numerical), then $i$-th value in such order is assigned $(i-0.5)/n$ normalized value. If there are multiple identical values, then they are all assigned the central position: $(i_{max}+i_{min}-1)/2n$ - allowing to also work with discrete values.

Each plot in Fig. \ref{pairs} contains such $n=1767$ points of coordinates being normalized variables. For discrete values, also in partially continuous variables (e.g. missing for Highest\_energy), we can lines in such plots. It is imperfect for discrete variables, but allows for some initial evaluation, visualization of dependencies - especially if combining with estimated joint distribution visualized in this Figure.
\begin{figure}[t!]
    \centering
        \includegraphics[width=8.5cm]{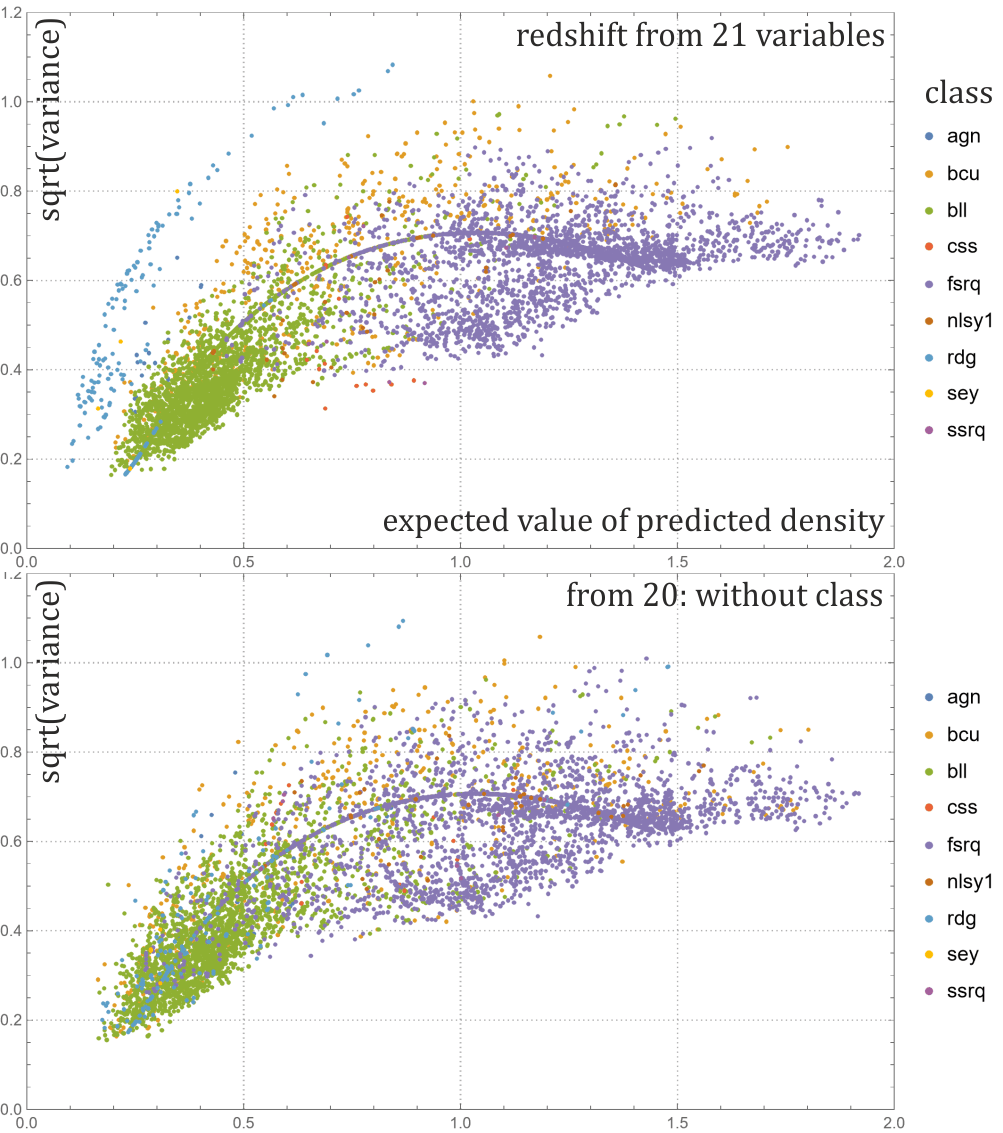}
        \caption{Expected values of predicted densities (vertical) vs square roots of their variances, with CLASS shown as color. The difference between them is removing class variable in the bottom plot - what reduces log-likelihood by $\approx 0.04$ (novelty of CLASS), e.g. through clear difference for rdg (light blue). We can see some strange curve in the center, which is yet to be interpreted. Generally we can see that different classes have different behaviour - it might be worth to consider separate models for them.  }
        \label{mv}
\end{figure}

\subsection{HCR density prediction}
Hierarchical Correlation Analysis (HCR)~\cite{hcr0} approach means working on (mixed) moment-like parameters chosen to allow to reconstruct the (joint) distribution from them. Conveniently for normalized variables, we represent (joint) density as a linear combination, here conditional density of predicted $y$ variable, based on the context (here $x=(x_1,..,x_{21}))$:
\be\tilde{\rho}(Y=y|X=x)=1+\sum_{i=1}^m f_i(y) a_i(x) \label{e1}\ee
Using ortonormal basis  of functions: $\int_0^1 f_i(x) f_j(x) dx = \delta_{ij}$, the $a_i$ parameters can be independently estimated as just mean of $f_i$ over the dataset. Here we would like to estimate them based on the context $x$, the mean is the value minimizing mean squared error from dataset, what suggests to just use MSE prediction of $a_i$ from $x$~(\cite{hcr1,hcr2}) - it could be done by some neural networks to be explored in the future, but for interpretability let us focus now on linear regression models: MSE predicts $a_i$ as a linear combination of features of $x$.

While we could use arbitrarily large orthonormal basis $m$ of e.g. Legendre polynomials, it can easily lead to overfitting, hence it is crucial to use e.g. cross-validation, here 10-fold, to make various types of decisions.

Linear combination (\ref{e1}) can get below zero, what is not allowed for density - hence, as in Fig. \ref{intr}, there is later used calibration: $\rho=\phi(\tilde{\rho})/Z$, using some function e.g. $\phi(\rho)=\max(\rho,0.1)$, here a bit better softmax-like: $\phi(\rho)=\ln(1+\exp(3\rho)/4)/3$. The $Z$ is normalization to integrate to 1. This integration is numerically costly, hence in practice we calculate values on regular lattice (size 1000 here) - approximating density with locally constant, this way replacing integration with summation.

\begin{figure}[t!]
    \centering
        \includegraphics[width=8.5cm]{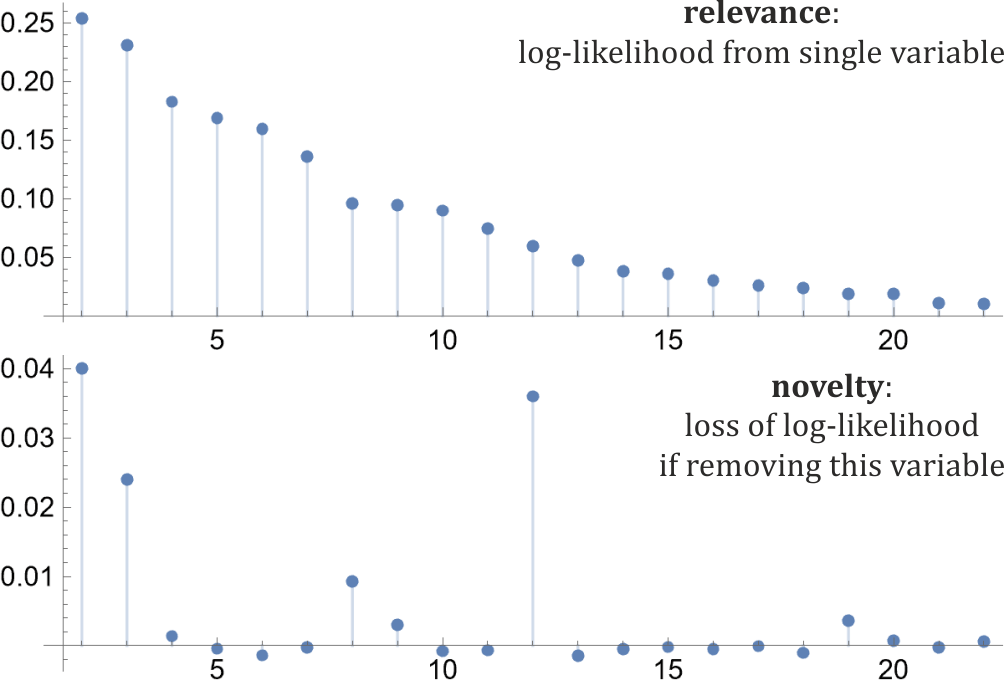}
        \caption{Evaluation of variables as in~\cite{hcr1} - \textbf{relevance} as log-likelihood for prediction from this single variable (estimated conditional entropy $-H(Y|X_j)$), it was used to choose their order. In contrast, \textbf{novelty} evaluates unique information available only from this variable ($H(Y|X_1..X_{j-1} X_{j+1}..X_k)-H(Y|X)$) - we can see the most valuable are 2 (CLASS), 12 (nuFun\_syn), 3 (nu\_syn), 8 (Counterpart\_Catalog), 9 (Highest\_energy) and 19 (Unc\_Counterpart). The remaining have individual contributions close to zero, or even negative. Trying to remove them and some other variables, the highest log-likelihood in cross-validation was 0.403 for just 11 variables marked in Fig. \ref{vars}: (2, 3, 4, 7, 8, 9, 12, 15, 19, 20, 22). Hence regularization can still be improved, what will require further research. }
        \label{rn}
\end{figure}

\subsection{Feature selection and CCA optimization}
As such features of $x$, for discrete values it is natural to start with one-hot encoding (vectors with single '1'), and for continuous we can use above orthonormal basis $(f_i(x_j))$.

There are also variables which are mostly continuous, but have discrete part - e.g. missing value. In such cases there were used both: $f_i(x_j)$ in the continuous part, and additional single one-hot vector: being '1' only in the single discrete value, and zero for the $f_i(x_j)$ features.

This way e.g. using up to $m=10$ degree polynomial, we get approx 200 features for $x=(x_1,..,x_{k})$ variables ($k=21$). Predicting separately multiple $a_i$ from them, we can easily get to model sizes in thousands - larger than dataset, hence we need some feature selection, regularization to avoid overfitting.

The main methodological contribution of this article is proposing CCA, explained in Appendix, for optimization of such features. It inexpensively and automatically optimizes linear subspace of strongest correlations - allows to choose optimized features being linear combinations of the original ones, which should contain nearly all dependencies.

Choosing maximal considered polynomial degree $m$ (generally $m$ could vary between variables), we predict $m$ features of $x$: $(a_1,\ldots,a_m)$, from $(f_1(x_j),\ldots,f_m(x_j))$ features of continuous variables, or "+1" for continuous-discrete here, or as the number of values for discrete variables.

CCA allows to find the most correlated linear combinations of such variables. Here it was used for $(f_1(y),\ldots,f_m(y))$ and all $\approx 21m$ features of $x$ to find the basis for predictions. For individual variables it was chosen differently: there were used separate CCA between $(f_1(y),\ldots,f_m(y))$ and features of this single variable - leading to basis presented in Fig. \ref{vars}.

It might be also worth to consider CCA optimized features directly e.g. for pairs (or larger numbers) of variables like $(f_{i_1}(x_{j_1}) f_{i_2}(x_{j_2}))$ - it allows to exploit also multivariate dependencies. However, this way the size of the model can rapidly growth, requiring subtle regularization techniques - some initial tests have allowed to improve log-likelihood by $\approx 0.01$ this way, what is planned to be examined further.

\subsection{Linear regression with l1 regularization}
Having CCA optimized basis for the predicted variable, and features of context $x$ - presented in Fig. \ref{vars}, we can use linear regression - with found coefficients presented in Fig. \ref{lasso}.

The final evaluations: log-likelihood with 10-fold CV, is improved if using regularization - here l1 "lasso": adding $\lambda |\sum_k c_k|$ to the minimized MSE cost, for $\lambda=50$ chosen to get nearly the highest evaluation. For equal treatment, the features are normalized before this linear regression - by subtracting the mean and dividing by square root of the variance.

This regularization has advantage of leading to sparse models - with low numbers of nonzero coefficients, also allowing for better interpretation. It is tempting to use it to discard the zero coefficients features based on the entire dataset, but it turned out to lead to overfitting. Previously used alternative approach~\cite{hcr2} is considering succeeding features, e.g. sorted by some relevance, and including them if  evaluation is improved - however, naively done it leads to overfitting, which might be avoided if using some acceptance threshold e.g. based on hypothesis testing.

This feature selection + regularization is very difficult to do right, will require further research, maybe using different techniques. In Fig. \ref{mv} we can see that even discussed careful optimization can be improved by just discarding 10 out of 21 used variables - the remaining likely still contain additional valuable information, but its practical exploitation requires careful further model optimization.

\section{Conclusions and further work}
While the standard of data science is still just prediction of values, there are many examples like the discussed, where available information does not allow to well localize the value - allowing only to predict probability distribution.

This version of article is rather methodological, to be improved through collaboration with experts in astronomy/astrophysics e.g. for choice of used variables, also for improved MSE predictions of $(a_i)$ parameters.

Then applications can range from initial knowledge for redshift estimation, understanding of subtle statistical dependencies, choice of exceptional objects for more detailed analyzis/observations/reexamination, maybe some improvements in classification.

There is also required further methodological research, especially for improved feature selectrion/regularization - to extract as much as possible of useful information, avoiding overfitting. For example just removing some variables as in Fig. \ref{rn}, log-likelihood in cross-validation was improved by $\approx 0.009$. From the other side, adding products of all pairs of first features $f_1(x_{j_1})f_1 (x_{j_2)}$ has allowed to improve log-likelihood by $\approx 0.01$, but it means rapid growth of featrues - requiring carefulness not to overfit.  This is a difficult optimization problem requiring further research.

\appendix
\textbf{Canonical Correlation Analysis} (CCA)~\cite{CCA} technique inexpensively finds strongly correlated linear subspaces for multidimensional random variables $X$, $Y$: we search for direction pairs $(a,b)$ maximizing correlation:
$$\operatorname*{argmax}_{a,b}\ \textrm{corr}(a^T X, b^T Y)$$
Applied multiple times, it leads to orthonormal set of vectors for $X$ and $Y$ - we can treat as features for prediction.

In practice it is calculated by whitening the variables - multiplication by ($C^{-1/2}$) matrix to get normalized variables of unitary covariance matrix, then perform SVD (singular value decomposition) of cross-covariance matrix for such normalized variables.

Specifically, for $\mu_X=E[X], \mu_Y=E[Y]$ expected values vectors, we need covariance and cross-covariance matrices:
\begin{small}
$$C_{XX}=E[(X-\mu_X)(X-\mu_X)^T],\quad C_{YY}=E[(Y-\mu_Y)(Y-\mu_Y)^T]$$
\end{small}
$$C_{XY}=E[(X-\mu_X)(Y-\mu_Y)^T]\qquad\quad C_{YX}=C_{XY}^T $$
Performing singular value decomposition (SVD), and returning to the original variables, we get
$$a\textrm{ is an eigenvector of }C_{XX}^{-1} C_{XY} C_{YY}^{-1} C_{YX} $$
\be b\textrm{ is proportional to } C_{YY}^{-1} C_{YX} a\ee
In practice we use some number of such vector pairs corresponding to the highest eigenvalues: strongest dependencies.

\bibliographystyle{IEEEtran}
\bibliography{cites1}
\end{document}